\documentclass[journal,twoside,web]{ieeecolor}
\usepackage{tmi}
\usepackage{cite}
\usepackage{amsmath,amssymb,amsfonts}
\usepackage{algorithmic}
\usepackage{graphicx}
\usepackage{textcomp}
\usepackage{hyperref}

\def\BibTeX{{\rm B\kern-.05em{\sc i\kern-.025em b}\kern-.08em
    T\kern-.1667em\lower.7ex\hbox{E}\kern-.125emX}}
\begin{document}
\title{AdaptiveSAM: Towards Efficient Tuning of SAM for Surgical Scene Segmentation }
%\title{AdaptiveSAM: A Hands-Free Method to make SAM Application Specific}
\author{Jay N. Paranjape, \IEEEmembership{Student Member, IEEE}, Nithin Gopalakrishnan Nair, \IEEEmembership{Student Member, IEEE},  Shameema Sikder, S. Swaroop Vedula, and Vishal M. Patel, \IEEEmembership{Senior Member, IEEE}
\thanks{    This research was supported by a grant from the National Institutes of Health, USA; R01EY033065. The content is solely the responsibility of the authors and does not necessarily represent the official views of the National Institutes of Health.}
\thanks{J. N. Paranjape, N. G. Nair, and V. M. Patel are with the Department of Electrical and Computer Engineering at Johns Hopkins University. (e-mail:\{jparanj1,ngopala2,vpatel36\}@jhu.edu).}
\thanks{S. Sikder is with the Wilmer Eye Institute, Johns Hopkins University School of Medicine, Baltimore, MD.(e-mail: ssikder1@jhmi.edu)}
\thanks{S. S. Vedula is with the Malone Center for Engineering in Healthcare, Johns Hopkins University.(e-mail: swaroop@jhu.edu)}
}
\maketitle

\begin{abstract}
Segmentation is a fundamental problem in  surgical scene analysis using artificial intelligence. However, the inherent data scarcity in this domain makes it challenging to adapt traditional segmentation techniques for this task. To tackle this issue, current research employs pretrained models and finetunes them on the given data. Even so, these require training deep networks with millions of parameters every time new data becomes available. A recently published foundation model, Segment-Anything (SAM), generalizes well to a large variety of natural images, hence tackling this challenge to a reasonable extent. However, SAM does not generalize well to the medical domain as is without utilizing a large amount of compute resources for fine-tuning and using task-specific prompts. Moreover, these prompts are in the form of bounding-boxes or foreground/background points that need to be annotated explicitly for every image, making this solution increasingly tedious with higher data size. In this work, we propose AdaptiveSAM - an adaptive modification of SAM that can adjust to new datasets quickly and efficiently, while enabling text-prompted segmentation. For finetuning AdaptiveSAM, we propose an approach called bias-tuning that requires a significantly smaller number of trainable parameters than SAM (less than 2\%). At the same time, AdaptiveSAM requires negligible expert intervention since it uses free-form text as prompt and can segment the object of interest with just the label name as prompt. Our experiments show that AdaptiveSAM outperforms current state-of-the-art methods on various medical imaging datasets including surgery, ultrasound and X-ray. Code is available at \href{https://github.com/JayParanjape/biastuning}{https://github.com/JayParanjape/biastuning}
\end{abstract}


\begin{figure*}[htp!]
\centering
  \includegraphics[width=\linewidth]{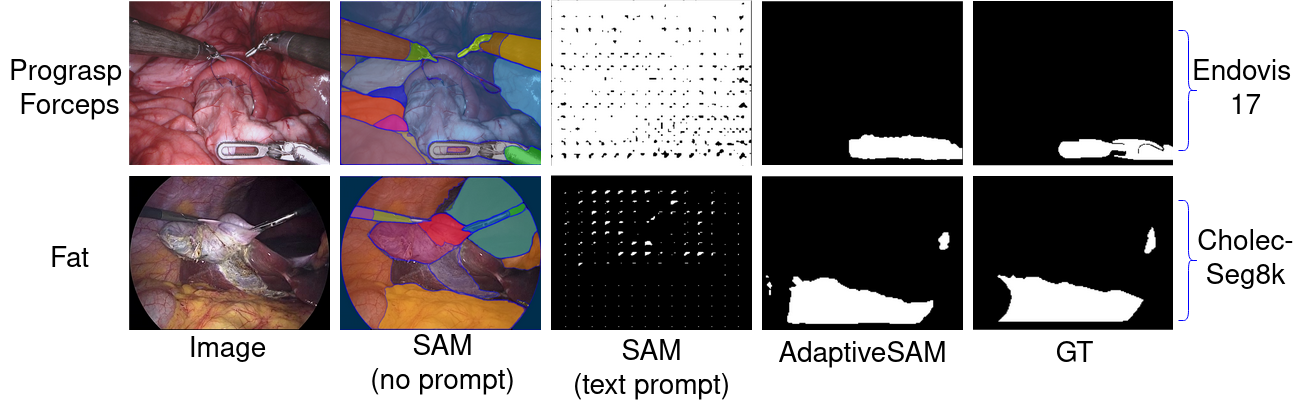} 
  \caption{Comparison of our method(AdaptiveSAM) with SAM. The text on the left denotes the prompt given to SAM and AdaptiveSAM. The datasets are indicated on the right. SAM, without any text prompts, produces good segmentation masks. However, we do not have control over segmenting specific objects without a manual prompt. If we add a text prompt to SAM as shown, we do not get good results, primarily because SAM is not trained with medical data or terminology. Our method produces good masks which can be controlled using text prompts. This makes it easier to use for surgical scene segmentation without the need for points/bounding boxes which must be manually given by experts.}
  \label{intro_fig_sam}
\end{figure*}

\section{Introduction}
\label{sec:introduction}
\begin{figure*}[t!]
\centering
  \includegraphics[width=\linewidth]{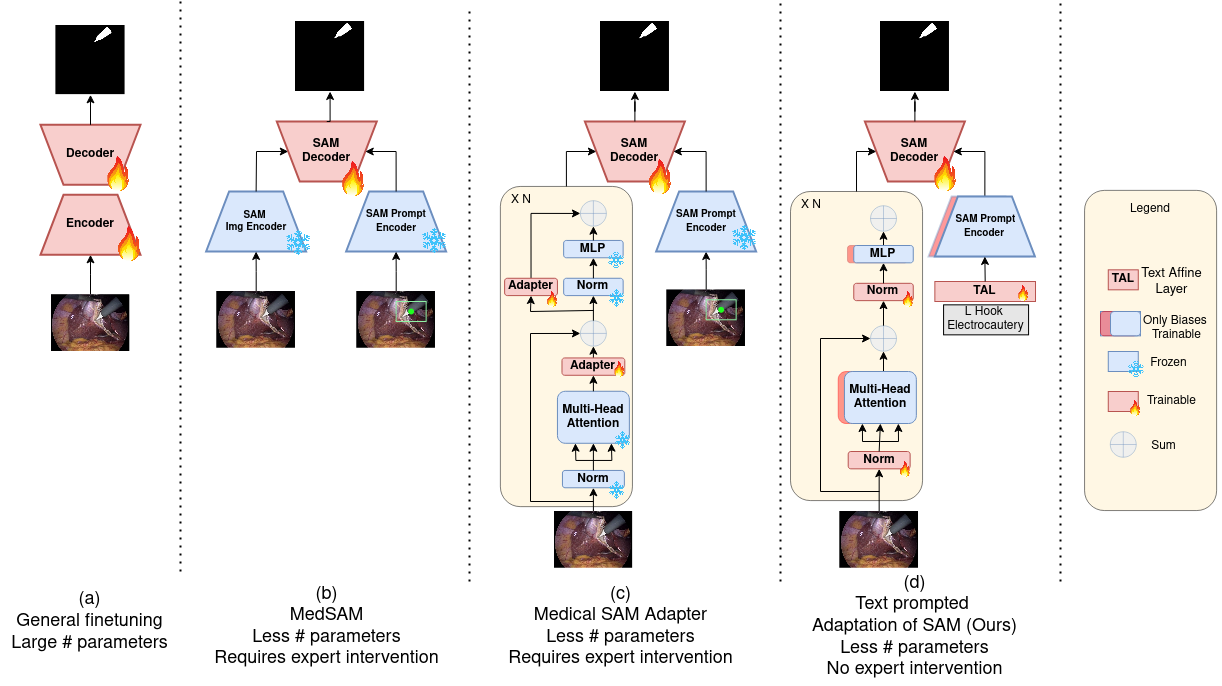} 
  \caption{Comparison of our method (d) with full finetuning(a) and other methods. (b) represents MedSAM\cite{ma2023segment}, which finetunes only the decoder part of SAM. (c) represents Medical SAM Adapter\cite{wu2023medical}, which also adds adapter layers inside the image encoder of SAM. Note that this method also adds similar adapter layers to the decoder instead of training all of its weights. This has not been shown in the figure for simplicity. Our method uses the SAM architecture as is but with trainable layer norms and biases. We also add a trainable Text Affine Layer(TAL) which transforms the text embeddings, helping SAM understand medical terminology.}
  \label{intro_fig}
\end{figure*}

Segmentation of surgical instruments, tissues and organs is an important task in analyzing and developing surgical systems. In human-guided surgeries, segmenting anatomical structures is useful for developing computer-assisted systems for surgeons\cite{Silva2022}. Semantic segmentation also plays a pivotal role in robotic surgery, where it is used for instrument tracking through the surgical process \cite{wang2023sam}. Deep learning-based solutions are widely adopted for tackling this problem the most prominent among them are UNet \cite{unet1} and its variants, which perform admirably well on surgical scene segmentation with proper training \cite{unetpp, nnunet, denseunet,ternaus}. However, these networks require substantial resources and time for every new dataset they are trained on. As shown in Figure \ref{intro_fig}, they generally comprise of an encoder and decoder that may have millions of parameters that need to be tuned for every new dataset. Hence, this limits the usage of these models. 

A similar problem also occurs in non-medical vision-based application scenarios, which is mitigated to some extent with the advent of foundational models. Foundation models are trained on a large corpus of images across the internet on multiple GPUs, which make them generalizable across different datasets. An example is CLIP \cite{clip} and its variants \cite{wang2022medclip,clip-variants} which are used for zero-shot visual recognition tasks. CLIP provides rich, generalized embeddings for images and text, which are closer to each other if the object related to the text appears in the image. Hence, it generates embeddings for similar objects closer in the latent space.
Similarly, Segment-Anything Model (SAM) was recently released as a foundational model for prompted segmentation \cite{sam}. Given a prompt in the form of a bounding box, points, mask or text, SAM can segment out the object that is related to the prompt. There have been multiple efforts to adapt SAM for different medical imagining-based applications by leveraging its generalized nature with minimal training \cite{hu2023efficiently,qiu2023learnable,wang2023sam,ma2023segment,wu2023medical}. We give a brief overview of these methods in Figure \ref{intro_fig}. While these finetuning-based methods reduce the training time and resources over general methods that train from scratch, the number of trainable parameters and the train time is still large due to inefficient freezing of SAM's layers or excessive architectural changes. To alleviate this, we propose a simple yet efficient strategy called bias-tuning, i.e. only tuning the bias parameters in the network. The effectiveness of bias has been widely studied in the Natural Language Processing (NLP) literature \cite{zaken2021bitfit} for fine-tuning large language models. Through our experiments and analysis, we found that tuning just the bias parameters and the normalization layers helps in effectively adapting a foundational model for surgical images. Another major issue with existing finetuning-based adaptation methods is that they include an additional constraint i.e. the need for the correct prompts. For the existing adaptation methods to work well with surgical datasets, expert intervention is needed to manually annotate the bounding boxes or points for the objects of interest. These prompts are required for all images during test time, making these approaches tedious to deploy when we have large data sizes.

In this work, we propose AdaptiveSAM - a solution that combines the best of both worlds. Just like general finetuning methods, it does not require expert intervention, and just like recent SAM-adaptation methods, it can be trained quickly and with fewer resources using bias-tuning. To achieve this, we introduce a trainable affine layer that can take as input a text prompt that is passed as input to the SAM model. We show this in Figure \ref{intro_fig}. The text prompt can simply be the label of interest and is required once for the entire dataset. Hence, our method does not require medical expertise unlike inputting per image bounding boxes as required by existing methods. As an example, if the label of interest is the 'Grasper', our method only expects the word itself as the input instead of a bounding box around the grasper or a foreground point on the grasper. While the latter has to be provided by an expert surgeon, our method does not need any expert intervention. AdaptiveSAM shows considerable improvements over SAM in terms of control as shown in Figure~\ref{intro_fig_sam}. SAM can be used without explicit bounding boxes or point prompts. However, as can be seen in Figure \ref{intro_fig_sam}, it cannot associate a mask with a given label without a prompt. AdaptiveSAM overcomes this limitation by allowing text prompts as inputs. To summarize, this paper makes the following contributions:

\begin{enumerate}
    \item We propose AdaptiveSAM - a text-prompted segmentation method for surgical datasets which requires no expert intervention and is able to produce highly precise results.
    \item We develop a training strategy called bias-tuning that can adapt foundational models like SAM for specific tasks with minimal training and resources. By applying bias-tuning for AdaptiveSAM, we only finetune 2\% of SAM's training parameters, making it trainable on a single GPU. 
    \item We achieve state-of-the-art (SOTA) results on three publicly available surgical scene segmentation datasets. We also show the generalizable nature of AdaptiveSAM with experiments on non-surgical datasets.
\end{enumerate}

\section{Related Work}
\subsection{Surgical Scene Segmentation}
Medical semantic segmentation recognizes and delineates different organs, tissues, or instruments in surgical images, enabling multiple substream tasks like tracking and classification. In general, medical image segmentation is challenging owing to various factors like similarity between different organs, spectral shadows cast by the operating instruments and so on \cite{Silva2022}. Various Convolutional Neural Network (CNN) based approaches have been proposed to tackle this task \cite{unet1,unetpp,nnunet,denseunet, maskrcnn, boundarymrcnn, isinet}. Ronneberger et al. proposed the UNet \cite{unet1} for this task which motivated further research into this field. Since then, various methods have been proposed which follow the structure of U-Net because of its simplicity and effectiveness \cite{unetpp, nnunet, denseunet,ternaus}. Other convolutional methods have also been proposed which first generate bounding boxes to identify the object of interest, followed by segmentation. Mask RCNN \cite{maskrcnn} and its variants \cite{boundarymrcnn,isinet} are notable models of this type. With the advent of transformers \cite{image_transformer}, various networks have been proposed that improve the global representation of CNNs for surgical scene segmentation \cite{trans1,trans2}. SAM uses Vision Transformers for segmentation as well.
However, all of the above approaches have a large number of parameters that need to be trained every time a new dataset is introduced. This limits the applicability of such methods to certain datasets or tasks. 

\subsection{Adaptation of SAM for medical applications}
A foundation model called SAM was recently released for the task of promptable segmentation \cite{sam}. SAM can segment images given prompts in the forms of bounding boxes, masks and points. However, SAM does not produce good results when used with medical images \cite{failsam1, failsam2, ma2023segment}. This is because the training data of SAM primarily consists of natural images. However, multiple works have adapted SAM \cite{ma2023segment,wu2023medical,hu2023efficiently} for medical segmentation with correct modifications and training. MedSAM \cite{ma2023segment} trains only the decoder part of SAM while keeping the encoders frozen. Medical SAM Adapter \cite{wu2023medical} also adds low rank adaption layers to the image and prompt encoders which can be trained while keeping the original SAM weights frozen. However, this method is memory intensive since these additional layers are added to every transformer block in the network which adds an additional computational overhead since the gradients need to be computed for these parameters and branches as well. AutoSAM \cite{hu2023efficiently} keeps the encoder frozen but adds a separate trainable predictor head that is CNN based, while \cite{qiu2023learnable} adds trainable CNN-based prompt layers after every transformer layer while keeping the SAM weights frozen. However, all of these methods are limited to using bounding boxes or points as prompts. Hence, if these models are to be deployed, experts would have to provide such prompts for every image or video frame, making them infeasible. In our work, we overcome this limitation by efficiently adapting text-based capabilities of SAM for medical data. 

\section{Proposed Method}
In this section, we introduce our approach for finetuning SAM for out-of-domain datasets. An overview of our approach can be found in Figure \ref{intro_fig} (d).
\subsection{Preliminaries: Architecture of SAM} 
Segment Anything Model(SAM) \cite{sam} proposes a new task of prompted segmentation where given an image and a prompt, the output is a mask specific to the prompt. The prompt can be in the form of a point, bounding box, mask or text. The prompt-based conditioning is achieved by having a separate image encoder and a prompt encoder, followed by a decoder that fuses the outputs of the image and prompt encoders. The image encoder of SAM has a Vision Transformer \cite{image_transformer} backbone and is pretrained with the Mask Auto-Encoder (MAE) strategy \cite{mae}. The prompt encoder supports four modalities for the prompts: point, bounding box, mask, and text. Point prompts are represented using positional embeddings and a foreground/background marker. Bounding box prompts are represented by the point embeddings of the top left and bottom right corners. Mask prompts are encoded using convolutions and text is encoded using CLIP pretraining \cite{clip}. The mask decoder is a lightweight transformer network, followed by a mask prediction head.

\subsection{Architectural Changes} 
Our goal is to adapt SAM to new surgical datasets for which SAM does not perform well. Finetuning based strategies\cite{ma2023segment, wang2022medclip, wu2023medical} have previously been found effective and are shown to work better than training from scratch because of their good initialization of weights. In our work, we wish to use the capability of SAM to capture rich discriminative features in the initial layers of the network. Hence, we use the same image encoder, prompt encoder and mask decoder as SAM and initialize their weights with the pretrained weights. Like SAM, text prompts are encoded using CLIP and images are passed to SAM's image encoder. However, CLIP fails to give meaningful text embeddings as CLIP is not trained with medical text terminology. Hence, to overcome this limitation we add a lightweight affine transformation layer that takes the pretrained CLIP embeddings as input and refines it to get medically relevant text embeddings as outputs. In Figure \ref{intro_fig} (d), this is represented by the Text Affine Layer (TAL). The operation of TAL is defined as follows 
\begin{equation}
    y = BatchNorm(ReLU(W_{TAL}^{T}X + b_{TAL})),
\label{tal}
\end{equation}
where \(X\) denotes the input CLIP embeddings for the given text, and \(y\) represents the transformed embeddings. \(W_{TAL}\) and \(b_{TAL}\) are the weights and biases of the Text Affine Layer. To summarize, for a given image-text pair, the image is encoded using SAM's image encoder, text is encoded using CLIP, followed by the TAL, and is fed into SAM's prompt encoder. These two outputs are fused in SAM's mask decoder, producing the output mask. We refer to the final model after applying these modifications as AdaptiveSAM.

\begin{figure*}[htp!]
\centering
  \includegraphics[width=\linewidth]{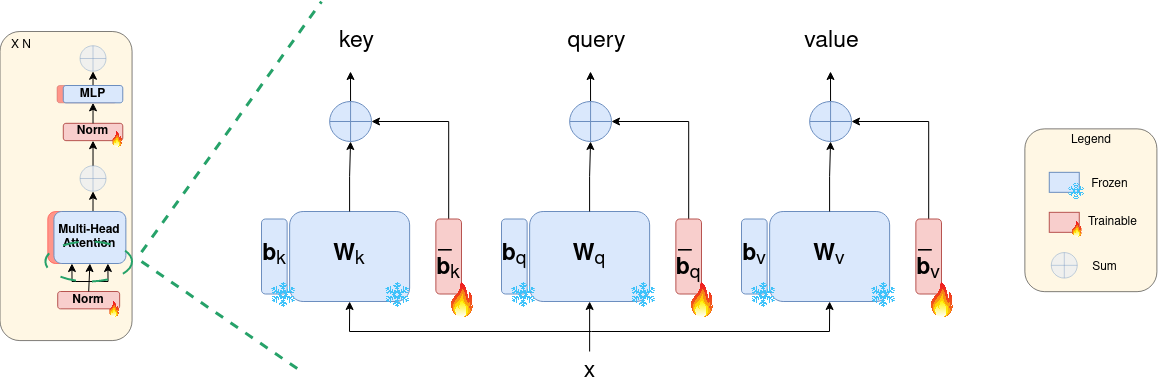} 
  \caption{An overview of bias-tuning. The first step of the attention module is the generation of keys, queries and values from the input. This step requires affine transformations as shown to the right. In bias-tuning, we keep all weights of SAM frozen. To account for the changed context from general images to surgical scene, we introduce trainable biases that are added to each affine transform output. This includes the multi-head attention as well as the MLP layer. Note that the latter is not shown in the figure explicitly for simplicity.}
  \label{model_arch}
\end{figure*}

\subsection{Training Strategy}
Existing approaches for adapting SAM for data-specific applications generally freeze the parameter-heavy encoders and finetune the lightweight decoder\cite{ma2023segment, hu2023efficiently}. Some approaches also try to perform low rank adaption by adding extra trainable layers in SAM's encoder\cite{wu2023medical}. In our training method, we propose a simpler yet effective method for finetuning SAM, called bias-tuning. As shown in Figure \ref{model_arch}, the image encoder of SAM has \(N\) blocks, each of which consists of a self attention block, followed by a Layer Norm and an MLP block, with skip connections. Given an image, it is divided into patches and passed to the self attention block. Here, each patch undergoes an affine transformation as described as follows
\begin{equation}
    qkv = (W_{qkv}^n)^Tx + b_{qkv}^n.
\label{qkv}
\end{equation}
Here, \(x\) represents the input image patch, \(W\) and \(b\) represent the weights and biases of the corresponding affine transformation, respectively and $n$ is used to index the layers in the encoder. The output is separated into three parts and are termed as query \((q)\), key \((k)\) and value \((v)\), which are used in the subsequent attention operation.

Note that during training such models, the weight matrices \(W_{qkv}\)s and their respective gradients are responsible for the majority of GPU memory usage. However, SAM is already trained with 1 Billion masks and hence can inherently segment general objects with its weight matrices. The task then, is to adapt it to the new context of surgical scenes, which can be achieved by adding a trainable shift parameter \(\Bar{b}\) to the outputs of each affine transformation. This is described in Eq. \eqref{qkv2} and \eqref{mlp2} as follows
\begin{equation}
    qkv = (W_{qkv}^n)^Tx + b_k^n + \Bar{b}_{qkv}^n,
\label{qkv2}
\end{equation}

\begin{equation}
    o_n = (W_{MLP}^n)^Tx + b_{MLP}^n + \Bar{b}_{MLP}^n.
\label{mlp2}
\end{equation}
Here, the output of a block is denoted by \(o_n\). During training, all other weights and biases are frozen and only the shift parameters are trainable, as shown in Figure \ref{model_arch}. These are initialized with zeros and trained with a small learning rate. We call this approach bias-tuning because we are essentially modifying only the biases of the encoder.
% This concept was inspired by a recent approach in <Nithin's paper reference and aim>, where similar scaling and shifting operations could transfer context information for diffusion models.

In addition, we also train the positional embeddings to facilitate segmentation on multiple scales as well as the Layer Norm layers since they tend to be biased towards SAM's training dataset.
Following MedSAM \cite{ma2023segment}, we let the decoder be fully trainable to learn how to fuse medical text prompts and surgical image embeddings correctly. Overall, the number of trainable parameters amounts to less than 2\% of SAM's trainable parameters.

Furthermore, in the surgical and medical domains, it is essential to build highly precise models with low false positive rates. In other words, if the text is 'Forceps' and there is no forceps present in the image, the output should be a blank mask. This property is lost if the model is solely trained on image-text pairs that only have non-zero masks, as is typically done in general-vision tasks. Hence, we also train the model with 'blank labels'. For a given image in a dataset, suppose there are \(N\) possible objects of interest. Then, for every image \(I\), we generate \(N\) image-text pairs irrespective of whether the object is present in the ground truth or not. We keep the ground truth mask empty for the latter case. We observe that performing training in this manner gives a boost in the ability of the model to generate correct (completely empty) masks when the object is indeed not present in the image. However, the approach introduces a significant data imbalance during training due to the total amount of empty regions in the training set ground truth. To mitigate this effect, we employ the focal loss defined as 
\begin{equation}\label{focal2}
    \mathbb{L}_{focal} = \frac{1}{B}\sum_{i=1}^{B}\sum_{j=1}^{HW} \mathbb{L}_{ij},
\end{equation}
where 
 \begin{equation}\label{focal1}
     \mathbb{L}_{ij} = 
     \begin{cases}
    -\alpha(1-p)^{\gamma}\log(p),& \text{if } y = 1\\
    -(1-\alpha)(p)^{\gamma}\log(1-p),              & \text{if } y = 0.\\
\end{cases}
 \end{equation}
Here, \(\mathbb{L}_{ij}\) denotes the loss corresponding to pixel $j$ in the \(i^{th}\) image, 
\(p \) corresponds to the probability of the prediction being 1, and
\(y \) denotes the true label of the pixel.
Note that \(B\) represents the batch size, \(H\) represents the height, and \(W\) represents the width of the mask. The hyperparameters \(\alpha\) and \(\gamma\) vary among datasets and need to be set through validation. As can be seen from the above equations, this loss significantly reduces the weight given to trivial predictions during training, thereby preventing the model from always producing blank masks.

\section{Experiments and Results}
We evaluate the proposed method on three publicly available surgical scene segmentation datasets - Endovis17 \cite{ev17}, Endovis18 \cite{ev18} and Cholec-Seg8k \cite{cholecseg8k}. For each of these datasets, we calculate the instrument-wise Dice scores for comparison. Following \cite{aiman}, we define the Dice Score (DSC) as follows
\begin{equation}
\label{dsc}
DSC = 
\begin{cases}
    \frac{2*|Y \cap \hat{Y}|}{|Y| + |\hat{Y}|},& \text{if } (|Y| + |\hat{Y}|) \neq 0\\
    1,              & \text{otherwise}.
\end{cases}
\end{equation}
We also calculate the Intersection over Union (IoU) Metric for comparison which is defined as 
\begin{equation}
\label{iou}
IoU = 
\begin{cases}
    \frac{|Y \cap \hat{Y}|}{|Y \cup \hat{Y}|},& \text{if } (Y \cup \hat{Y}) \neq \phi\\
    1,              & \text{otherwise}
\end{cases}
\end{equation}
where \(Y\) represents the true segmentation map and \(\hat{Y}\) represents the predicted segmentation map. In what follows, we describe the implementation details and the results corresponding to each of these datasets. In addition, we show the generalization capability of our method on non-surgical scenes by conducting experiments on ultrasound and X-ray imaging datasets. 

\subsection{Experimental Setup}
In our experiments, we initialize the AdaptiveSAM image encoder with the pretrained ViT-base model for SAM. We augment the training images with random rotations up to an angle of 10 degrees, random saturation and brightness changes with scale factor 2. For Endovis 17 and Endovis 18, we set hyperparameters \(\alpha = 0.75\) and \(\gamma = 3\). For the Cholec-Seg8k dataset, we set these as \(\alpha = 0.5\) and \(\gamma = 0\). For all datasets. we train with a batch size of 32 and training is performed on a single Nvidia Quadro RTX 8000 GPU and requires around 12 GB of memory. We use the AdamW optimizer with a fixed learning rate of 1e-4 for all cases.

\subsection{Datasets}
The Endovis 17 (EV17) dataset \cite{ev17} contains eight videos for training and ten videos for testing from the Da-Vinci robotic system. We use two videos out of the training videos for validation. This dataset has labels for six robotic instruments - Grasping Forceps, Bipolar Forceps, Large Needle Driver, Grasping Retractor, and Monopolar Curved Scissors. The Endovis 18( EV18) dataset \cite{ev18} is a robotic instrument clinical dataset that includes organs and surgical items. It has sixteen sequences for training and four sequences for testing. We use four sequences from the training set for validation. This dataset has annotations for ten objects including different organs, tissue and surgical instruments. Finally, the Cholec-Seg8k dataset \cite{cholecseg8k} has annotations for twelve objects including organs, tissues, and two surgical instruments. We follow the train, validation and test splits as described in \cite{Silva2022}.

\subsection{Results}
\noindent \textbf{Results on Endovis 17 (EV17).}
Results corresponding to EV17 are tabulated in Table \ref{ev17}. This dataset has a total of ten test sets. We average the classwise Dice scores and IoU scores across all ten test datasets. We compare the performance of our method with recent 
 convolutional and transformer-based state-of-the-art segmentation methods \cite{unet1, transunet, medT}. We also use SAM with text prompt without any training (called Zero Shot performance of SAM or SAM-ZS) \cite{sam} for comparison. As can be seen from Table \cite{ev17}, UNet \cite{unet1} and TransUNet \cite{transunet} perform poorly on EV17. This is because of the severe class imbalance present in the dataset and convolutional methods like UNet are unable to tackle this due to limitations in the training loss which doesn't deal with class imbalances. TransUnet has a transformer-based encoder which requires a greater amount of data to produce good embeddings. However, EV17 has various classes that are only present in a few frames in a single sequence. This affects the training process for the method. Med-T \cite{medT} was introduced to overcome this weakness of requiring more data points. However, we observe that for EV17, Med-T tends towards producing blank masks even when the object is present in the image, showing its dependence on the label distribution of the data. In contrast, AdaptiveSAM makes use of the highly generalized nature of SAM to produce good results with low amount of data and training. We also observe a significant jump in performance over the zero-shot performance of SAM (68\% jump in DSC), showcasing the effectiveness of our adaptation technique for this dataset.\\

\begin{figure*}[htp!]
\centering
  \includegraphics[width=\linewidth]{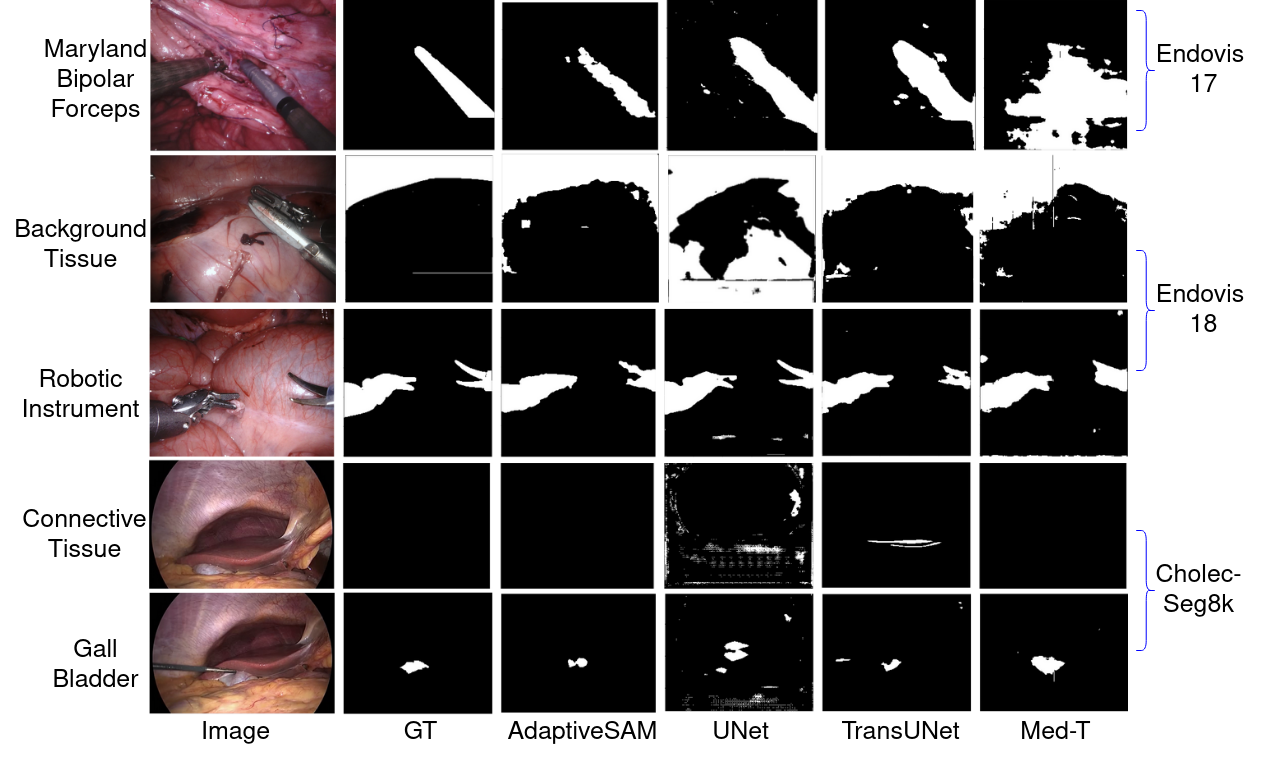} 
  \caption{Qualitative performance of AdaptiveSAM and other methods on different datasets and text prompts. From the top, the input text prompts given to AdaptiveSAM are `Monopolar Curved Scissors', `Background Tissue', `Robotic Instrument', `Connective Tissue', and `Gall Bladder' respectively as indicated on the left. The first row is from EV17, the next two rows are from EV18 and the last two rows are from Cholec-Seg8K.}
  \label{preds}
\end{figure*}

\noindent \textbf{Results on Endovis 18 (EV18). } Results corresponding to EV18 are tabulated in Table \ref{ev18}. The dataset has 4 sequences for testing. We generate results for all of them for every object and report the average object-wise DICE Scores and IoU scores across the four test datasets. Among the current state-of-the-art methods, \cite{groupev18} proposes an interesting approach of grouping objects and calculating the groupwise metrics. For the 10 objects in the datasets, they propose grouping them into five classes - Organs, Robotic parts, Suturing, Other Tools and Background. Hence, in Table \ref{ev18}, we report the groupwise scores as well. In addition, we compare our approach with the winning method for Endovis 18 \cite{ev18} challenge. Please note that all of these are methods that trained all the network parameters. AdaptiveSAM achieves significantly higher DSC and IoU scores over these methods, with an improvement of 8\% in DSC and 14\% in IoU Score. We also observe a significant improvement of 21\% in the DSC over zero shot performance of SAM.\\ 

\noindent \textbf{Results on Cholec-Seg8k: } Results corresponding to Cholec-Seg8k are shown in Table \ref{cholec}. We follow the splits from previous research for reporting our scores \cite{Silva2022}. Among the twelve classes of objects present in this dataset, there is limited representation of five classes - Blood, Hepatic Vein, Connective Tissue, Liver Ligament and Cystic Duct. Hence, these were grouped into one Miscellaneous category in previous research \cite{Silva2022}. However, we keep these as separate entities so as to allow our Text Affine Module to distinguish between these categories and the background. On an average across all classes, we see an improvement of 2\% over the SOTA method as seen in the table. Furthermore, AdaptiveSAM improves the Zero Shot performance of SAM by 60\% in DSC.

Please note that we do not include the results from MedSAM \cite{ma2023segment} and MedSAMAdapter \cite{wu2023medical} since these methods require providing a point or bounding box input per image which must be manually done. Hence, it would not be fair to compare against methods that do not require intervention. We also observe that SAM-ZS does poorly in all the 3 datasets. SAM-ZS uses CLIP to generate prompt embeddings from the given text. However, CLIP is trained on general datasets and hence, we see that SAM-ZS does not perform well when used as is. AdaptiveSAM learns a lightweight transformation over the frozen CLIP embeddings that enables it to train well. SAM-ZS serves as a baseline as well as an ablation for AdaptiveSAM, thus showing the effectiveness of our method over zero-shot performance of SAM. For all of the above datasets, we observe that AdaptiveSAM significantly improves upon this. As mentioned earlier, AdaptiveSAM is able to use the generalized weights learnt by SAM while successfully shifting its biases to adapt to the data shift observed between natural and surgical scene images. Hence, with less amount of training time and resources, it is able to adapt SAM to perform better than existing SOTA methods.\\

\noindent \textbf{Qualitative Evaluation of Results}
Qualitative results corresponding to our experiments on surgical scene segmentation are shown in Figure \ref{preds}. As can be seen from this figure, AdaptiveSAM is capable of producing highly precise masks for different types of text prompts. For other methods that do not require text prompts, the models output segmentation maps for each of the classes. We follow the following equation 
\begin{equation}
    \text{final mask}_i = 
\begin{cases}
    1, & \text{if } \textit{argmax(M, dim=0)} = i\\
    0,              & \text{otherwise}
\end{cases}
\label{finalmask}
\end{equation}
to get the final class-wise mask for comparison. Here, \(i\) represents the class index and \(M\) represents the output of the segmentation model. \(M\) is assumed to have a shape of \(CXHXW\), where \(C\) is the number of classes, \(H\) is the height and \(W\) is the width of the mask. AdaptiveSAM will correctly output a blank mask when a particular object is absent in the image. Furthermore, in comparison to other methods, it produces less noisy outputs. This can be attributed to the property of identifying complete 'objects' exhibited by the original SAM model. This property is also retained by AdativeSAM which allows it to produce closed masks with less amount of noise.\\

\begin{figure}[htp!]
\centering
  \includegraphics[width=.45\linewidth]{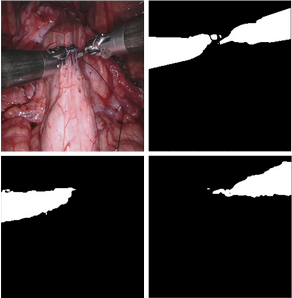} \;\;\; \includegraphics[width=.45\linewidth]{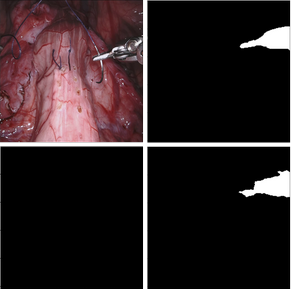}\\
  (a) \hskip110pt(b)
  \caption{Location learning capability of AdaptiveSAM. From the top left, in clockwise order, image, ground truth, prediction with the text prompt ``Right Large Needle Driver" and prediction with the text prompt ``Left Large Needle Driver". (b) From the top left, in clockwise order, image, ground truth, prediction with the text prompt ``Right Large Needle Driver" and prediction with the text prompt ``Left Large Needle Driver". Note that since there is no needle driver to the left, the corresponding mask is blank.}
  \label{fig:spatial1n2}
\end{figure}

\noindent \textbf{Location Learning Capabilities.}
AdaptiveSAM can also learn spatial correspondences between the text and the image. As shown in Figure \ref{fig:spatial1n2} (a), there are two large needle drivers present in the image. With the input text as ``Left Large Needle Driver", AdaptiveSAM only segments out the needle driver on the left side of the image. Similarly, as shown in Figure \ref{fig:spatial1n2} (b), only the right needle driver is present in the image. Hence, for the text prompt ``Left Large Needle Driver", a blank mask is returned. This capability can be learnt with the correct annotations during training and is made possible with the use of freeform text.

% We encourage future work to exploit this property of AdaptiveSAM further to facilitate tracking applications and segmentation using more complex text prompts.

%\begin{figure}[hbt!]
%\centerline{\includegraphics[width=0.4\columnwidth]{images/spatial1.png}}
%\caption{Location learning capability of AdaptiveSAM. From the top left, in clockwise order, image, ground truth, prediction with the text prompt ``Right Large Needle Driver" and prediction with the text prompt ``Left Large Needle Driver"}
%\label{spatial1}
%\end{figure}

%\begin{figure}[!thb]
%\centerline{\includegraphics[width=0.4\columnwidth]{images/spatial2.png}}
%\caption{Location learning Capability of AdaptiveSAM. From the top left, in clockwise order, image, ground truth, prediction with the text prompt ``Right Large Needle Driver" and prediction with the text prompt ``Left Large Needle Driver". Note that since there is no needle driver to the left, the corresponding mask is blank.}
%\label{spatial2}
%\end{figure}

\subsection{Results on Non Surgical Datasets}
\begin{figure*}[htp!]
\centering
  \includegraphics[width=\linewidth]{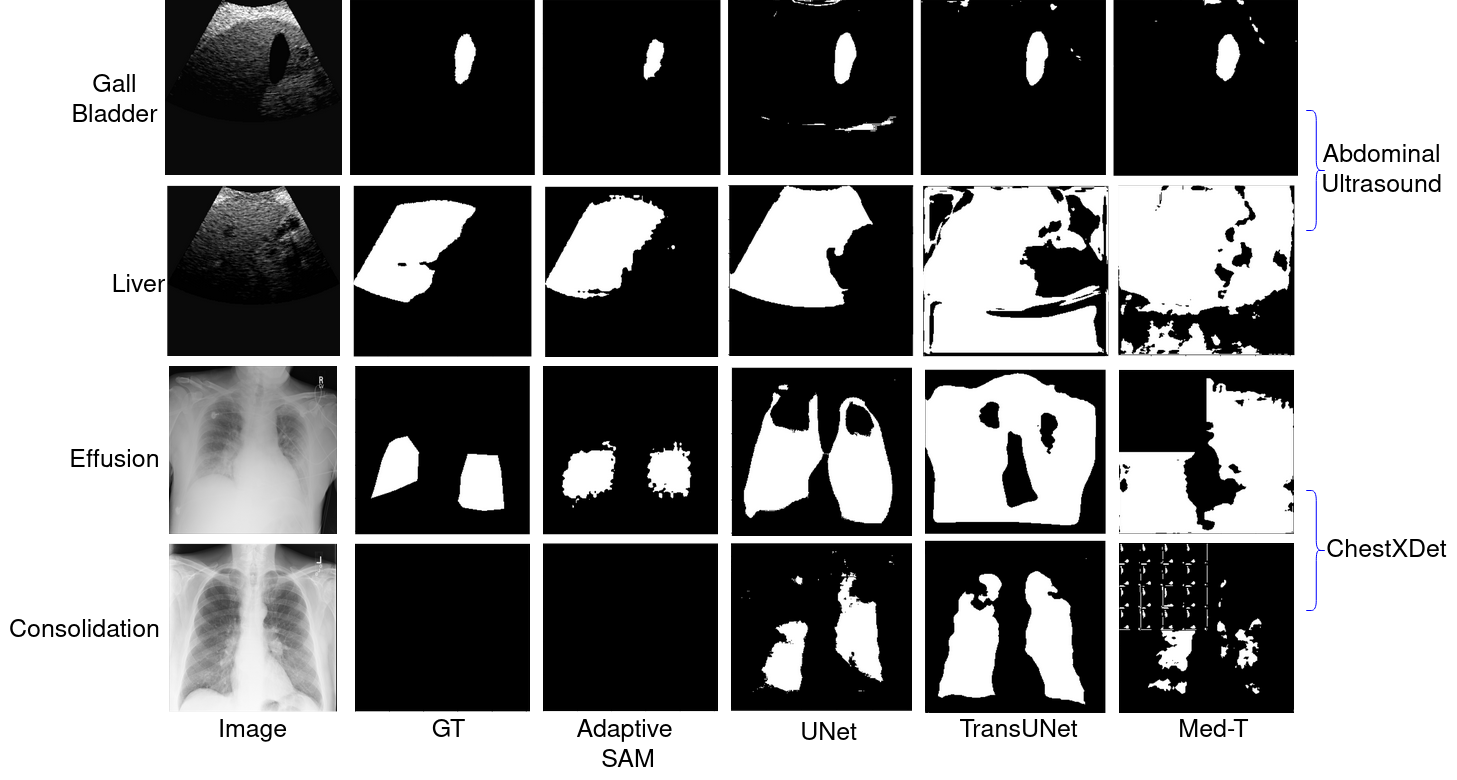} 
  \caption{Qualitative performance of AdaptiveSAM and other methods on the Abdominal Ultrasound dataset \cite{ultrasound} and ChestXDet \cite{chestxdet}, with different text prompts. The input text prompts given to AdaptiveSAM are `Gall Bladder', `Liver', `Effusion' and `Consolidation' respectively as indicated on the left. The first two rows are from Ultrasound Dataset and the next two rows are from ChestXDet.}
  \label{us_egs}
\end{figure*}

 To show the significance of our approach on non-surgical datasets, we apply AdaptiveSAM on the Abdominal Ultrasound Dataset \cite{ultrasound} for the task of object segmentation. This dataset has real and synthetic images of abdominal ultrasounds with segmentation masks available for eight classes - Liver, Kidney, Pancreas, Vessels, Adrenals, Gallbladder, Bones and Spleen. While training data only consists of simulated ultrasound images, the test data has both real and synthetic data. We get an average DSC of 0.48 on the synthetic test set and 0.58 on the real test set, which is significantly greater than other methods, as shown in Table \ref{ultrasound}. Some qualitative results are shown in Figure \ref{us_egs} Rows 1 and 2. Note that ultrasound imaging is a separate modality from surgical images, thus leading to a significant domain shift in the images, even though the labels are similar. However, AdaptiveSAM is able to perform well on these images as well, producing precise and less noisy masks. 
 
We also test our method on one more modality, namely X-ray images. We use the ChestXDet Dataset \cite{chestxdet}, which has X-ray images and annotations for thirteen classes - Effusion, Nodule, Cardiomegaly, Fibrosis, Consolidation, Emphysema, Mass, Calcification, Pleural Thickening, Pneumothorax, Fracture, Atelectasis and Diffuse Node. We observe that AdaptiveSAM significantly outperforms other methods as shown in Table \ref{xray} since it can handle imbalance in the dataset better than other methods. Figure \ref{us_egs} Rows 3 and 4 show some qualitative results on this dataset.

\begin{table*}
\begin{center}
\centering
\caption{Results on Endovis17. PF - Prograsp Forceps, BF - Bipolar Forceps, LND - Large Needle Driver, GR - Grasping Retractor, VS - Vessel Sealer, MCS - Monopolar Curved Scissors, DSC - Dice Score, IoU - Intersection over Union Score. Some of the results were taken from respective papers directly and hence, all metrics were not available. These are represented by "-". }
% \label{table}
\setlength{\tabcolsep}{3pt}
% \begin{tabular}{|p{25pt}|p{25pt}|p{25pt}|}
\begin{tabular}
{|c|c|c|c|c|c|c|c|}
\hline
Method &
\multicolumn{7}{c|}{Object wise DSC[IoU] }\\
\hline
& PF & BF & LND & GR & VS & MCS & Avg. \\
\hline
UNet & 0.03[0.02] & 0.07[0.04] & 0.28[0.22] & 0.19[0.19] & 0.02[0.02] & 0.08[0.06] & 0.11[0.09] \\
TransUNet & 0.08[0.05] & 0.10[0.02] & 0.20[0.14] & 0[0] & 0.03[0.02] & 0.10[0.05] & 0.08[0.05] \\
MedT & 0.29[0.28] & 0.21[0.19] & 0.31[0.27] & 0.61[0.61] & 0.36[0.36] & 0.06[0.04] & 0.31[0.29] \\ 
S3Net\cite{baby2023forks} & - [0.54] & - [0.75] & - [0.62] & - [0.27] & - [0.36] & - [0.43] & - [0.50] \\
TraSeTR\cite{zhao2022trasetr} & - [0.57] & - [0.45] & - [0.56] & - [0.11] & - [0.39] & - [0.31] & - [0.34] \\
Mask2Former\cite{cheng2022maskedattention} & - [0.20] & - [0.20] & - [0.45] & - [0] & - [0.12] & - [0.01] & - [0.14] \\
SAM-ZS & 0.03[0.01] & 0.04[0.02] & 0.08[0.05] & 0[0] & 0.07[0.04] & 0.11[0.06] & 0.06[0.03]
\\
AdaptiveSAM(Ours) & 0.64[0.60] & 0.54[0.52] & 0.71[0.67] & 0.91[0.91] & 0.81[0.80] & 0.82[0.81] & \textbf{0.74}[\textbf{0.72}] \\
\hline
\end{tabular}
\label{ev17}
\end{center}
\end{table*}

\begin{table*}
\centering
\caption{Results on Endovis 18. Bg Tissue - Background Tissue, RI - Robotic Instrument, KP - Kidney Parennchyma, CK - Covered Kidney, SN - Suturing Needle, Suction Ins. - Suction Instrument, S. Intestine - Small Intestine, UP - Ultrasound Probe, DSC - Dice Score, IoU - Intersection over Union Score. Some of the results were taken from respective papers directly and hence, all metrics were not available. These are represented by "-". Some methods group certain labels into one category and report the groupwise results, denoted by multi columns numbers in the table.}
% \label{table}
\setlength{\tabcolsep}{3pt}
% \begin{tabular}{|p{25pt}|p{25pt}|p{25pt}|}
\begin{tabular}
% {|p{66pt}|p{33pt}|p{33pt}|p{33pt}|p{33pt}|p{33pt}|p{33pt}|p{33pt}|p{33pt}|p{33pt}|p{33pt}|p{33pt}|}
{|c|c|c|c|c|c|c|c|c|c|c|c|}
\hline
Method &
\multicolumn{11}{c|}{Object wise Dice Score(DSC) [IoU Score]}\\
\hline
 & Bg Tissue & RI & KP & CK & S. Intestine & Thread & SN & Clamps & Suction Ins. & UP & Avg. \\
\hline
LinkNet34\cite{linknet} & 0.77[0.63] & 0.87[0.78] & \multicolumn{3}{c|}{0.23[0.14]} & \multicolumn{2}{c|}{0.74[0.59]} & \multicolumn{3}{c|}{0.33[0.20]} & 0.59[0.47]
\\
LinkNet50\cite{linknet} & 0.76[0.61] & 0.87[0.78] & \multicolumn{3}{c|}{0.21[0.12]} & \multicolumn{2}{c|}{0.73[0.57]} & \multicolumn{3}{c|}{0.37[0.24]} & 0.59[0.46]
\\
\cite{groupev18} & 0.78[0.64] & 0.88[0.79] & \multicolumn{3}{c|}{0.26[0.16]} & \multicolumn{2}{c|}{0.75[0.61]} & \multicolumn{3}{c|}{0.37[0.24]} & 0.61[0.49]\\
Winner, EV18\cite{ev18} & - & -[0.70] & -[0.65] & -[0.28] & -[0.34] & -[0.48] & - & -[0.85] & - & -[0.28] & -[0.51]
\\
SAM-ZS & 0.41[0.26] & 0.17[0.10] & 0.25[0.16] & 0.08[0.04] & 0[0] & 0[0] & 0[0] & 0[0] & 0.05[0.02] & 0.01[0.01] & 0.10[0.06]
\\
UNet & 0.64[0.49] & 0.74[0.66] & 0.34[0.23] & 0.16[0.13] & 0.9[0.9] & 0.01[0] & 0.91[0.91] & 0.72[0.72] & 0.36[0.34] & 0.04[0.03] & 0.48[0.36] \\
TransUNet & 0.73[0.59] & 0.70[0.60] & 0.49[0.32] & 0.33[0.33] & 0.63[0.63] & 0.01[0.01] & 0.04[0.04] & 1[1] & 0.66[0.64] & 0.62[0.61] & 0.52[0.48] \\
MedT & 0.56[0.40] & 0.66[0.54] & 0.26[0.16] & 0.44[0.44] & 0.78[0.78] & 0.82[0.82] & 0.90[0.90] & 0.96[0.96] & 0.62[0.61] & 0.84[0.84] & 0.68[0.64] \\
AdaptiveSAM(Ours) & 0.66[0.51] & 0.68[0.57] & 0.31[0.21] & 0.33[0.33] & 0.57[0.55] & 0.88[0.88] & 0.91[0.91] & 0.82[0.82] & 0.86[0.86] & 0.85[0.85] & \textbf{0.69[0.65]} \\
\hline
\end{tabular}
\label{ev18}
\end{table*}

\begin{table*}
\begin{center}

\caption{Results on Choec8k. GB - Gall Bladder, AW - Abdominal Wall, GT - Gastrointestinal Tract, CD - Cystic Duct, LHEC - L Hook Electrocautery, HV - Hepatic Vein, CT - Connective Tissue, LL - Liver Ligament, DSC - Dice Score, IoU - Intersection over Union Score. Some of the results were taken from respective papers directly and hence, IoU metric was not available. Hence, they are not shown in the table for certain rows. Some methods group certain labels into one category and report the groupwise results, denoted by multi columns numbers in the table.}
% \label{table}

\setlength{\tabcolsep}{3pt}
% \begin{tabular}{|p{25pt}|p{25pt}|p{25pt}|}
\scalebox{0.8}{
\begin{tabular}{|c|c|c|c|c|c|c|c|c|c|c|c|c|c|}
\hline
Method &
\multicolumn{13}{c|}{Object wise DSC[IoU] }\\
\hline
& Fat & Liver & GB & AW & GT & Grasper & LHEC & Blood & HV & CT & LL & CD & Avg. \\
\hline
U-Net\cite{unet1} & 0.87 & 0.52 & 0.40 & 0.73 & 0.26 & \multicolumn{2}{c|}{0.52} & \multicolumn{5}{c|}{0.08} & 0.48 \\

U-Net++\cite{unetpp} & 0.91 & 0.75 & 0.63 & 0.83 & 0.11 & \multicolumn{2}{c|}{0.61} & \multicolumn{5}{c|}{0.16} & 0.62 \\

DynUnet\cite{nnunet} & 0.89 & 0.82 & 0.57 & 0.84 & 0.13 & \multicolumn{2}{c|}{0.57} & \multicolumn{5}{c|}{0} & 0.55 \\

UNetR\cite{trans2} & 0.88 & 0.74 & 0.42 & 0.76 & 0.35 &\multicolumn{2}{c|}{0.71} & \multicolumn{5}{c|}{0} & 0.55 \\

DeepLabV3+\cite{deeplabv3p} & 0.86 & 0.74 & 0.60 & 0.81 & 0.14 &\multicolumn{2}{c|}{0.62} & \multicolumn{5}{c|}{0.12} & 0.56 \\

% SAM-ZS & 0.05 & 0 & 0.02 & 0 & 0 & 0.01 & 0.04 & 0.01 & 0.14 & 0.01 & 0.14 & 0.01 & 0.04
% \\
% & [0.02] & [0] & [0.02] & [0] & [0] & [0.01] & [0.04] & [0.01] & [0.14] & [0.01] & [0.14] & [0.01] & [0.03]
% \\
SAM-ZS & 0.05[0.02] & 0[0] & 0.02[0.02] & 0[0] & 0[0] & 0.01[0.01] & 0.04[0.04] & 0.01[0.01] & 0.14[0.14] & 0.01[0.01] & 0.14[0.14] & 0.01[0.01] & 0.04[0.03]
\\
% TransUNet & 0.83 & 0.43 & 0.77 & 0.35 & 0.43 & 0.70 & 0.55 & 0.61 & 0.82 & 0.57 & 0.72 & 0.64 & 0.62 \\
% &[0.72] & [0.28] & [0.65] & [0.22] & [0.33] & [0.60] & [0.52] & [0.61] & [0.80] & [0.53] & [0.72] & [0.64] & [0.55] \\

TransUNet & 0.83[0.72] & 0.43[0.28] & 0.77[0.65] & 0.35[0.22] & 0.43[0.33] & 0.70[0.60] & 0.55[0.52] & 0.61[0.61] & 0.82[0.80] & 0.57[0.53] & 0.72[0.72] & 0.64[0.64] & 0.62[0.55] \\

% MedT & 0.81 & 0.39 & 0.56 & 0.34 & 0.25 & 0.48 & 0.71 & 1 & 0.70 & 0.69 & 0 & 0.89 & 0.57 \\
% & [0.69] & [0.25] & [0.41] & [0.21] & [0.17] & [0.35] & [0.65] & [1] & [0.70] & [0.62] & [0] & [0.89] & [0.50] \\

MedT & 0.81[0.69] & 0.39[0.25] & 0.56[0.41] & 0.34[0.21] & 0.25[0.17] & 0.48[0.35] & 0.71[0.65] & 1[1] & 0.70[0.70] & 0.69[0.62] & 0[0] & 0.89[0.89] & 0.57[0.50] \\

% AdaptiveSAM(Ours) & 0.85 & 0.71 & 0.37 & 0.80 & 0.10 & 0.20 & 0.70 & 1 & 0.70 & 0.38 & 1 & 1 & \textbf{0.64} \\
% & [0.75] & [0.55] & [0.25] & [0.68] & [0.10] & [0.15] & [0.63] & [1] & [0.70] & [0.38] & [1] & [1] & \textbf{[0.60]} \\

AdaptiveSAM(Ours) & 0.85[0.75] & 0.71[0.55] & 0.37[0.25] & 0.80[0.68] & 0.10[0.10] & 0.20[0.15] & 0.70[0.63] & 1[1] & 0.70[0.70] & 0.38[0.38] & 1[1] & 1[1] & \textbf{0.64[0.60]} \\
\hline
\end{tabular}}

\label{cholec}
\end{center}
\end{table*}

\begin{table*}
\begin{center}
\centering
\caption{Results on the Ultrasound dataset. SAM-ZS denotes zero-shot performance of the original SAM on the dataset. DSC - Dice Score, IoU - Intersection over Union Score}
% \label{table}
\setlength{\tabcolsep}{3pt}
% \begin{tabular}{|p{25pt}|p{25pt}|p{25pt}|}
\begin{tabular}
% {|p{66pt}|p{33pt}|p{33pt}|p{33pt}|p{33pt}|p{33pt}|p{43pt}|p{33pt}|p{33pt}|p{33pt}|}
{|c|c|c|c|c|c|c|c|c|c|}
\hline
Method &
\multicolumn{9}{c|}{Objectwise DSC[IoU] }\\
\hline
& Liver & Kidney & Pancreas & Vessels & Adrenals & \footnotesize{Gall Bladder} & Bones & Spleen & Avg. \\
\hline
SAM-ZS & 0.17[0.17] & 0.20[0.20] & 0.72[0.72] & 0.21[0.21] & 0.44[0.44] & 0.65[0.65] & 0.67[0.67] & 0.63[0.63] & 0.46[0.46]\\

UNet & 0.28[0.20] & 0.37[0.35] & 0.11[0.09] & 0.16[0.13] & 0.85[0.85] & 0.08[0.06] & 0.17[0.15] & 0.14[0.13] & 0.27[0.24]\\

TransUNet & 0.18[0.11] & 0.09[0.06] & 0.03[0.02] & 0.03[0.01] & 0[0] & 0.11[0.08] & 0.05[0.03] & 0.02[0.01] & 0.08[0.04]\\

MedT & 0.18[0.12] & 0.03[0.01] & 0.27[0.26] & 0.10[0.09] & 0.85[0.85] & 0.15[0.13] & 0.02[0.01] & 0.08[0.08] & 0.21[0.19]\\

AdaptiveSAM(Ours) & 0.36[0.30]& 0.30[0.29] & 0.50[0.50] & 0.40[0.40] & 0.86[0.86] & 0.63[0.63] & 0.67[0.66] &
0.54[0.54] &
\textbf{0.53}[\textbf{0.52}] \\
\hline
\end{tabular}
\label{ultrasound}
\end{center}
\end{table*}

\begin{table*}
\begin{center}
\centering
\caption{Results on the ChestXDet dataset. SAM-ZS denotes zero shot performance of the original SAM on the dataset. Ef - Effusion, No - Nodule, Cm - Cardiomegaly, Fb - Fibrosis, Co - Consolidation, Em - Emphysema, Ma - Mass, Ca - Calcification, Pt - Pleural Thickening, Pn - Pneumothorax, Fr - Fracture, At - Atelectasis, Dn - Diffuse Node, DSC - Dice Score, IoU - Intersection over Union Score}
% \label{table}
\setlength{\tabcolsep}{3pt}
\scalebox{0.8}{

\begin{tabular}
% {|p{66pt}|p{33pt}|p{33pt}|p{33pt}|p{33pt}|p{33pt}|p{33pt}|p{33pt}|p{33pt}|p{33pt}|p{33pt}|p{33pt}|p{33pt}|p{33pt}|p{33pt}|}
{|c|c|c|c|c|c|c|c|c|c|c|c|c|c|c|}
\hline
Method &
\multicolumn{14}{c|}{Object wise DSC[IoU] }\\
\hline
& Ef & No & Cm & Fb & Co & Em & Ma & Ca & Pt & Pn & Fr & At & Dn & Avg. \\
\hline
SAM-ZS & 0.05[0.04] & 0.13[0.13] & 0.53[0.53] & 0.36[0.36] & 0.15[0.15] & 0.28[0.28] & 0.23[0.23] & 0.10[0.10] & 0.37[0.37] & 0.07[0.07] & 0.40[0.40] & 0[0] & 0.26[0.26] & 0.22[0.22]\\ 
UNet & 0.15[0.08] & 0.08[0.08] & 0.06[0.04] & 0[0] & 0.13[0.09] & 0.02[0.01] & 0.95[0.95] & 0[0] & 0.08[0.08] & 0[0] & 0.50[0.50] & 0.02[0.02] & 0.02[0.01] & 0.15[0.14]\\
TransUNet & 0.06[0.04] & 0.87[0.87] & 0.06[0.04] & 0.59[0.59] & 0.13[0.08] & 0.01[0.01] & 0.89[0.89] & 0[0] & 0.74[0.74] & 0[0] & 0.08[0.08] & 0[0] & 0[0] & 0.26[0.25]\\
MedT & 0.06[0.03] & 0.75[0.75] & 0.08[0.07] & 0.01[0.01] & 0.10[0.06] & 0.03[0.02] & 0.12[0.12] & 0[0] & 0.91[0.91] & 0[0] & 0[0] & 0.37[0.37] & 0.07[0.07] & 0.19[0.19]\\
AdaptiveSAM(Ours) & 0.52[0.51] & 0.88[0.88] & 0.86[0.86] & 0.86[0.86] & 0.43[0.42] & 0.93[0.92] & 0.95[0.95] & 0.91[0.90] & 0.84[0.83] & 0.93[0.93] & 0.86[0.86] & 0.94[0.93] & 0.93[0.93] & \textbf{0.83[0.83]}\\
\hline
\end{tabular}}
\label{xray}
\end{center}
\end{table*}

\section{Conclusion}
In this paper, we introduced an adaptation of SAM that allows it to perform well on surgical data, called AdaptiveSAM. We also develop a novel and lightweight method called bias-tuning that allows efficient finetuning of large-scale models like SAM and make them more application specific with minimal training and resources. Further, we also introduce text-prompted segmentation for surgical datasets which addresses the problem of expert supervision required by other adaptation methods. AdaptiveSAM only requires the user to provide the object of interest once, using free-form text instead of specifying a foreground point/bounding box for every image, as required by existing methods. Hence, it is easier to use. Adding text as input also opens up the possibility of enhancing AdaptiveSAM to understand more complex queries in the future. We motivate this by showcasing its ability to identify spatial correspondences between text and image. We show that our method outperforms existing state-of-the-art methods on three widely used surgical datasets. Finally, we also show that bias-tuning and text prompted segmentation can be generalized to non-surgical datasets as well by comparing the performance of AdaptiveSAM with existing SOTA segmentation methods on Ultrasound and X-ray modalities.

\bibliographystyle{ieeetr}
\bibliography{biastuning_v4}

\end{document}